# Clustering based on the In-tree Graph Structure and Affinity Propagation

Teng Qiu
School of Life Science and Technology
University of Electronic Science and Technology of China
Chengdu, China
qiutengcool@163.com

Longjie Li
School of Life Science and Technology
University of Electronic Science and Technology of China
Chengdu, China
liyj@uestc.edu.cn

*Abstract*—A recently proposed clustering method, called the Nearest Descent (ND), can organize the whole dataset into a sparsely connected graph, called the In-tree. This ND-based In-tree structure proves able to reveal the clustering structure underlying the dataset, except one imperfect place, that is, there are some undesired edges in this In-tree which require to be removed. Here, we propose an effective way to automatically remove the undesired edges in In-tree via an effective combination of the In-tree structure with affinity propagation (AP). The key for the combination is to add edges between the reachable nodes in In-tree before using AP to remove the undesired edges. The experiments on both synthetic and real datasets demonstrate the effectiveness of the proposed method.

## I. INTRODUCTION

Clustering is an old and fundamental problem in multiple disciplines such as statistics, pattern recognition, and machine learning, aiming at dividing a set of data into groups based on the similarities or distances [17], [8]. It has a wide application in diverse fields ranging from science, engineering to business. With the rapid growth of the unlabeled data in this big data era, it becomes more and more uneconomical to label those raw data by hand. For this reason, clustering methods become more significant than ever. However, despite the fact that many clustering methods have been proposed, clustering still remains quite challenging and more effective clustering methods are always expected by scientists and engineers [8]. The difficulty is largely due to the unsupervised nature of this task.

In the following, we first briefly describe two existing methods, which are the basis of the proposed method in this paper.

### A. Nearest Descent (ND)

Recently, a physically inspired clustering method, called the Nearest Descent (ND) [14], was proposed by mimicking the evolutionary behavior of the complex particle system. Specifically, each data point in ND is viewed as a basic particle that could generate a negative isotropic field centered at it, and the fields from different data points are linearly superposed. Consequently, a non-uniform potential distribution is formed, in which the points in the denser point areas usually have lower potentials. And in turn, this non-uniform space would trigger the evolution of the point distribution, that is, data points tend to move from higher potential areas to lower ones (denoted as the "descending direction").

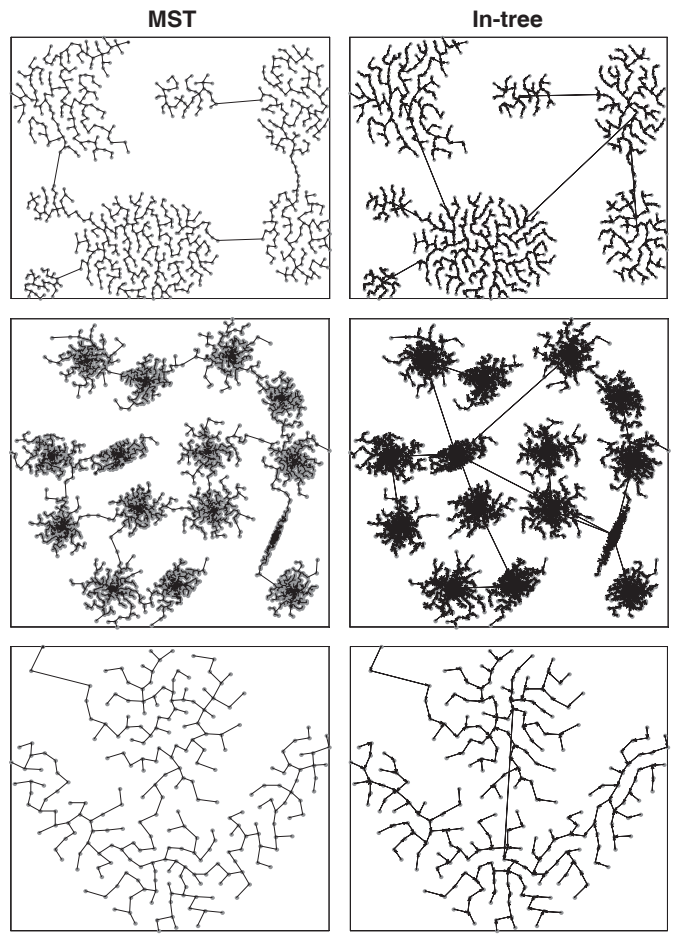

Fig. 1. Comparison between MST (left) and ND-based In-tree (right) on several datasets. The undesired edges between clusters in each In-tree are obviously more distinguishable (at least much longer) than the undesired edges in each MST.

Instead of seeking to the analytic solution for the trajectory of the moving path of each point, a simple rule is devised to mimic the moving process, that is, to let each point $i$ "descend" (referring to the descending direction) to the nearest neighbor point $I_i$. Consequently, the trajectory of each point can be approximated by a directed path consisting of a sequence of zigzag "hops" (i.e., directed edges), similar to the case in Isomap [15]. The descending direction constraint in the rule endows the constructed network with such beautiful order that each node has only one directed path to reach the node (actually the root node) with the globally lowest potential. As proven in [14], the constructed network is precisely the In-tree structure in graph theory. In-tree, also called in-arborescence or in-branching, is a digraph that meets: (i) only one vertex has outdegree 0; (ii) any other vertice has outdegree 1; (iii) there is no cycle in it; (iv) it is a connected graph.

The connections in this In-tree structure can well capture the underlying cluster structure in the data, being consistent with the distribution of the data points, except some undesired edges that need to be removed. However, as the comparison in Fig. 1 shows, unlike the undesired edges between clusters in Minimal Spanning Tree (MST) [18], those undesired edges in In-tree are very distinguishable and thus are easy to be determined [14]. Some effective edge removing methods have already been reported in [14], [13].

### B. Affinity propagation (AP)

In [5], Frey and Dueck proposed an effective clustering method called Affinity propagation (AP), which is able to automatically find a set of optimal data points called the exemplars (actually the cluster centers) from the dataset that can maximize the sum of the similarities for all the other points to their corresponding exemplars.

AP takes as input two kinds of real-valued data: (i) the similarity $s(i,j)$ of any pair of data points; (ii) a real number $s(i,i)$ for each node $i$. $s(i,j)$ signals how well node $j$ is suited to act as the exemplar of point $i$. $s(i,i)$, also called "preference", indicates how likely point $i$ is to be an exemplar, and the number of clusters can be automatically determined by $s(i,i)$. Since each point $i$ is equally suitable to be the exemplar, $s(i,i)$ is suggested to be the common value $\eta$. A large $\eta$ would lead to a small number of clusters, while a small $\eta$ would result in a large cluster number. Therefore, $\eta$ is suggested to be the median value of the similarities so as to produce a moderate number of clusters.

Based on the above inputs, AP uses a ***message-passing*** strategy to update the network until convergence. This process requires only simple and local computation, and is much more effective than $K$-means [11]. For instance, in certain dataset, one run of AP can be superior to more than 10000 runs of $K$-means clustering.

However, AP has two principle defects: (i) like $K$-means, AP is not able to detect the non-spherical clusters [10], [1], that is, when a cluster is of irregular shape, AP would divide it into multiple ones; (ii) AP is not established on a well-founded theory that can always guarantee its convergence or optimality [9]. We have seen several efforts made to tackle the first problem. For instance, in [10], a method called the soft-constrained affinity propagation (SCAP) was proposed by introducing a penalty term to the cost function, and in [1], a method called the Minimum Curvilinear affinity propagation (MCAP) was proposed by taking the negative value of the so-called "Minimum Curvilinear" distance (instead of the Euclidean distance) as the input of AP. Both SCAP and MCAP achieved better performances than AP in detecting certain irregular clusters in the data from biology.

## II. MOTIVATION, RATIONALE AND ADVANTAGE OF THE PROPOSED METHOD

Our motivation comes from the following two aspects. On the one hand, the methods to remove the undesired edges in the ND-based In-tree in [14], [13] are more of the interactive or semi-supervised ones, whereas in some cases (e.g., in cases where clustering only serves as a pre-processing step), users would prefer to get clustering result automatically without any form of supervision. On the other hand, interactive methods are largely relied on good enough intermediate graphs, which is hard to achieve when the dataset contains unexplicit cluster structure. For the above reasons, it is valuable to design automatic approach to remove the undesired edges in the In-tree.

In this paper, via an effective combination of In-tree with AP, an automatic approach is proposed to remove the undesired edges in the ND-based In-tree. The rationale for this combination is that the ND-based In-tree, together with the potential values on all nodes, actually contains all the inputs needed for AP: (i) the potential values can serve as a priori to specify the preference of each node as an exemplar; (ii) the weights of the directed edges can specify the likelihood of each node to be the exemplar of other nodes. Thus, AP could serve as an effective method to remove those undesired edges automatically.

Besides the merit that the proposed method provides an automatical In-tree-based clustering procedure, the proposed method has some other advantages. For instance, due to the merit of the density-based ND, the proposed method is able to detect the non-spherical clusters that AP cannot. Also, due to the sparseness and effectiveness of the graph, the message-passing procedure in AP would be much faster to converge in the proposed method, as compared with the case in which the message-passing procedure is run on the whole pair-wise similarity matrix of the dataset.

## III. DETAILS OF THE PROPOSED METHOD

Given a dataset of $N$ data points or nodes, the proposed clustering method consists of the following three steps:

**Step 1, construct the In-tree structure by ND.** First, let the potential $P_i$ at each node $i$ be the superposition of the isotropic Gaussian potentials (negative value) exerted from all points:

$$P_i = -\sum_{j=1}^{N} e^{-\frac{d_{i,j}^2}{\sigma}}, \qquad (1)$$

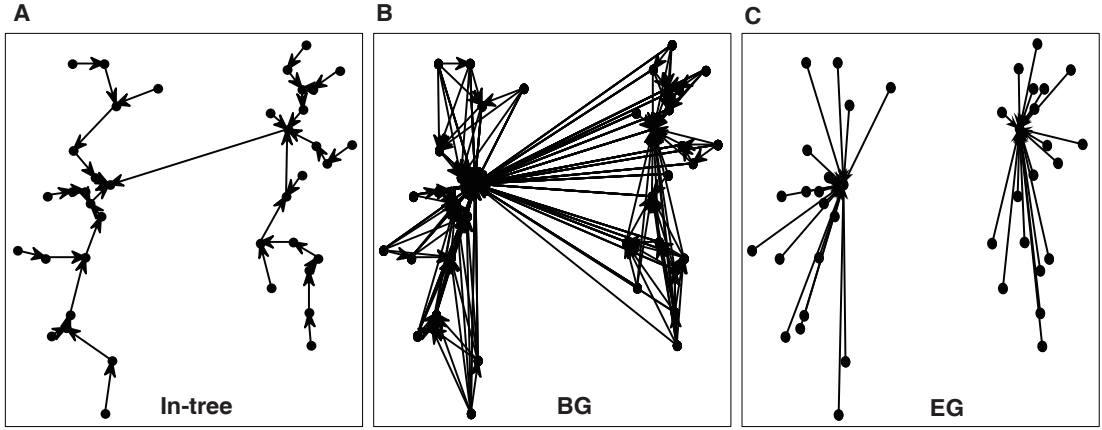

Fig. 2. An illustration of the proposed method for the graphs generated in each step. (A) The In-tree structure. (B) The belief graph. (C) The exemplar graph.

where $d_{i,j}$ denotes the pair-wise distance (e.g., Euclidean) between nodes $i$ and $j$, and $\sigma$ is the parameter with positive value. Then, according to the ND rule, each node $i$ "descends" to the nearest node, denoted as $I_i$, where the node $I_i$ is also called the parent node of node $i$, and "descends" means that the parent node $I_i$ should have lower potential than node $i$. Therefore, such parent node $I_i$ of any node $i$ can be defined as follows (i.e., the nearest node among the ones with lower potentials):

$$I_i = \arg\min_{j \in J_i} d_{i,j} \quad (2)$$

where $J_i = \{j | P_j < P_i\}$ denotes all the *candidate* parent nodes of node $i$. Note that this definition here for $I_i$ is a simple version. See a more elaborate definition in [14], by which the In-tree data structure can be guaranteed in any circumstance. In this In-tree structure, as shown in Fig. 2A, except one node (the node with the lowest potential), each other node $i$ and its parent node $I_i$ actually define one directed edge, and the distance between them defines the edge length or weight. The parent node is the end node of the directed edge.

**Step 2, construct the belief graph.** Any node $i$ is linked with node $j$ by a directed edge if node $i$ can reach node $j$ along a directed path $\Gamma_{i,j}$ in In-tree. This means that more additional directed edges are added to In-tree in this step, as shown in Fig. 2B. We call this extended graph the belief graph (**BG**). The edge weight $W_{BG}(i,j)$ is defined as the sum of the weights of all edges in $\Gamma_{i,j}$. When node $j$ is the parent node of node $i$, there is only one edge in $\Gamma_{i,j}$, that is, the directed edge between node $i$ and its parent node $I_i$.

Thanks to the In-tree structure, there is always only one directed path from any node $i$ to its reachable node $j$ (thus the end node $j$ is easy to be determined). This is in contrast to Isomap [15], in which a time-consuming process is required to search for an appropriate (e.g., the shortest) path among several choices in an undirected $K$ (requiring to be specified beforehand) nearest neighborhood (KNN) graph.

**Step 3, identify exemplars by AP.** The similarity between nodes $i$ and $j$ is defined as

$$s(i,j) = e^{-\frac{W_{BG}(i,j)}{\sigma}} \quad (3)$$

if the two nodes are connected by a directed edge in BG; otherwise $s(i,j) = 0$. The initial preference $s(i,i)$ at node $i$ is set proportional to the sum of the similarities between node $i$ with all the nodes that can reach it. Since the similarity between node $i$ and any other node is defined as zero, for simplicity, $s(i,i)$ can be written as

$$s(i,i) = \alpha \sum_{j=1}^{N} s(i,j) \quad (4)$$

where $\alpha$ is the parameter with negative value. Then, based on the input, the sparse version of AP is used and thus the messages are propagated only between the connected nodes (i.e., $s(i,j) \neq 0$).

After a couple of iterations (or the so-called messages-passing), the exemplars (or the cluster centers) will be determined. The directed edges between the non-exemplar nodes and their corresponding exemplars will be preserved, while other edges will be removed. We call the trimmed graph the exemplar graph (**EG**), as shown in Fig. 2C.

## IV. EXPERIMENTS

### A. Parameters

The proposed method contains two parameters: $\sigma$ in Eq. 1 and $\alpha$ in Eq. 4. Only $\alpha$ is an additional parameter introduced by the proposed method, since $\sigma$ is a parameter already existing in ND while computing the potentials on nodes with Eq. 1. And it has been shown in [14] that ND is not sensitive to $\sigma$ in a relatively wide range. Also, in the experiments here, we will show that the proposed method is not very sensitive to $\alpha$, either. An empirical value for $\sigma$ could be

$$\sigma = \sum_{i,j} d_{i,j} / N^2 / \log(N). \quad (5)$$

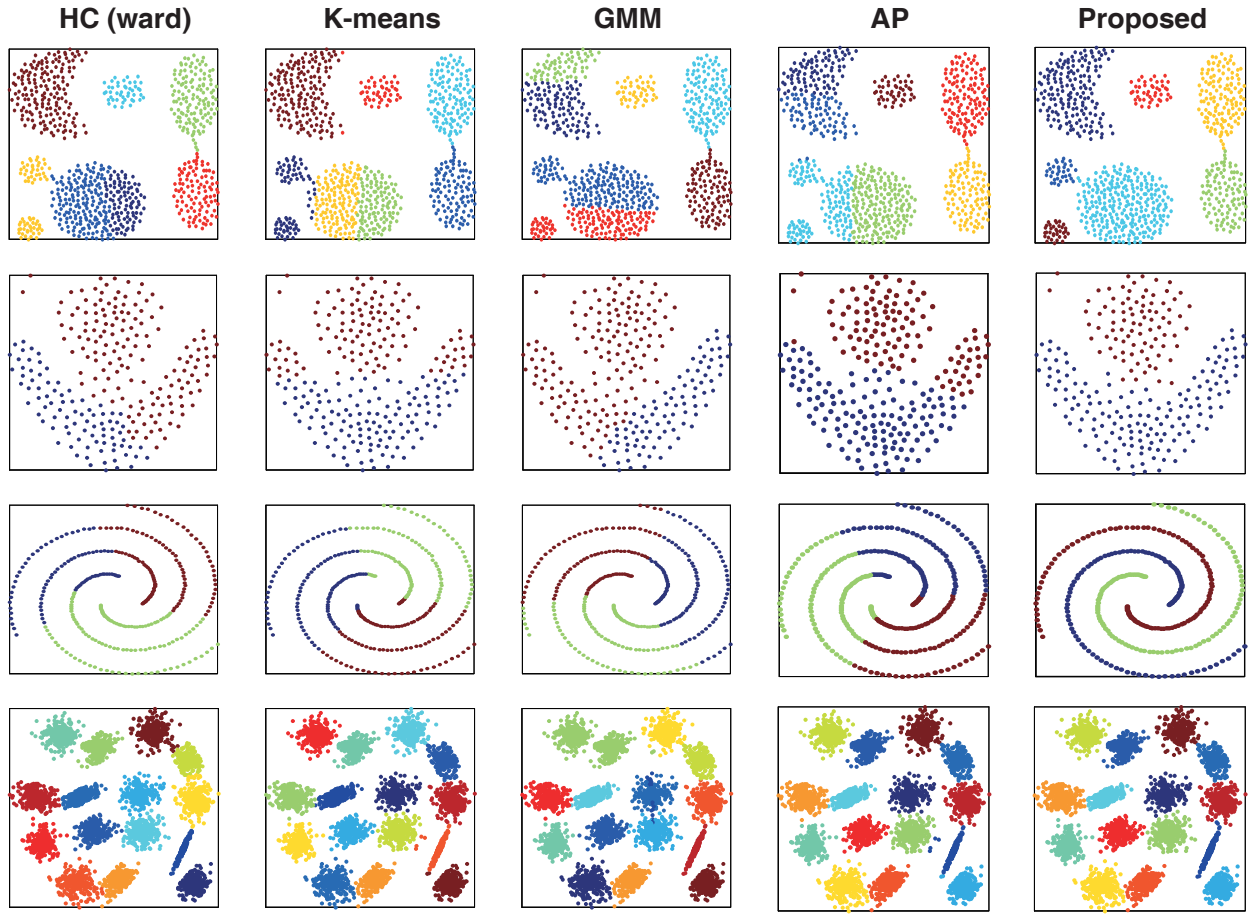

Fig. 3. Comparison between the proposed method (right-most) and other popular clustering methods. Only the proposed method can successfully detect the clusters in all datasets.

This definition is quite similar to the corresponding parameter defined in [9]. And an empirical value for $\alpha$ is suggested to be as follows

$$\alpha = -N/2. \quad (6)$$

These empirical values can serve as the references for further adjustment.

### B. Compared methods

Besides AP, we also compared the proposed method with other popular methods as K-means [11], Gaussian Mixture model (GMM) [12] and Hierarchical clustering (HC) (e.g., ward's method) [16]. For K-means and GMM, since they are sensitive to initialization, we run them 20 times and reported the best results. For HC, we cut the dendrogram to get the same cluster number as other methods on each dataset. For AP, we set the pair-wise similarities as negative values of the corresponding distances and we set the common preferences carefully so as to achieve the same cluster numbers as other methods. Note that, in the experiments, all the comparisons among different methods are under the same cluster numbers and same distance metric. For numerical datasets, the simple Euclidean distance is used.

### C. Evaluation index

When testing the high-dimensional datasets, we use *purity* ($p$) to show the consistency between the obtained cluster assignments and the ground truth, written as

$$p = \frac{1}{N} \sum_{j=1}^{M} \max_{k} |C_j \cap L_k| \quad (7)$$

where $C_j$ denotes the $j$-th generated cluster and $L_k$ denotes the $k$-th class provided by the dataset. It ranges from 0 to 1. A larger value of the purity indicates a better performance.

### D. Results

First, we tested the performance of the proposed method on four 2-dimensional synthetic datasets from [3], [7], [6], [2], which are usually hard for one clustering algorithm to successfully find the cluster structures in all datasets. As Fig. 3 shows, HC, K-means, GMM, and AP fails to discover the non-spherical cluster structures on the first three datasets. In contrast, the proposed method can accurately detect all the true clusters, indicating that (i) the proposed method achieves the goal of automatically removing the undesired edges in the In-trees in Fig. 1 and (ii) the combination of the ND-based

In-tree and AP in the proposed method well solves the defect of AP in detecting non-spherical clusters [10], [1].

In the above experiments, we set the parameter $\sigma = 1$, 1 and 1 (around the empirical values defined by Eq. 5) for the first three datasets and set $\sigma$ precisely as the empirical value for the last dataset. The other parameter $\alpha$ were set as the empirical values with Eq. 6 for all datasets. Besides, further experiment show that a large range of values for $\alpha$ can lead to the same results for the proposed method. For example, for the second dataset, the magnitude of $\alpha$ can vary from 4 to 4600.

Then, we tested the proposed method on five high dimensional synthetic datasets[1] from [4]. All these datasets contain $N = 1024$ points sampled from $M = 16$ different multivariate Gaussian functions with the dimensionality varying from 32, 64, 256, 512, to 1024. Using the empirical values for both $\sigma$ and $\alpha$ according to Eqs. 5 and 6, the proposed method achieved on each dataset the perfect performance of 16 clusters without error clustering assignment. Nevertheless, these datasets are not so easy for all other methods. For instance, for the 512-dimensional (D-512) dataset, only AP and HC achieved the same performance as the proposed method. The performance of K-means and GMM are much poorer, as Table I shows.

As Table I shows, we also tested the proposed method on one of the most popular real-world datasets, the Iris dataset[2] which contains 150 samples, each represented by 4 real-valued features. On this dataset, the proposed method achieved a purity of 0.953 with 3 clusters, superior to other methods. On another real-world dataset, the Banknote dataset[3] with 1372 samples and 4 real-valued features, the proposed method also achieved a performance much better than other methods in the case of 7 clusters.

In addition, in order to further compare the proposed method and AP, we tested them on the widely used Mushrooms dataset[4] (Fig. 4A). This real-world dataset collects 8124 mushrooms, each featured by characters rather the real values. As Fig. 4A shows, each row represents a mushroom consisting of 22 characters. Numerical methods such as K-means and GMM are not able to handle such kind of data. In contrast, since the proposed method only takes as input the similarity or distance of pair-wise data points, it can handle data of any type. Here, the distance $d_{i,j}$ between any pair of mushroom $X_i$ and $X_j$ is measured by $\sum_m 1\{X_i^m \neq X_j^m\}$, where $1\{X_i^m \neq X_j^m\}$ equals 1 if the $m$-th attributes are different, else 0. It has been shown in [14] that the In-tree structure for this dataset is insensitive to a large range value of the parameter $\sigma$ from 0.001 to 1000. Here, we chose $\sigma = 4$ for instance and tested the proposed method on the Mushroom dataset with the other parameter $\alpha$ varying in a large range. As Fig. 4B shows, the performance of the proposed method is overall quite satisfactory, which is actually comparable with that of other methods as surveyed in [17]. Compared with AP, the proposed method is superior in clustering result (Fig. 4C) and requires much less time to converge (Fig. 4D) due to the sparse feature of the input.

[1]From http://cs.joensuu.fi/sipu/datasets/
[2]From http://archive.ics.uci.edu/ml/datasets/Iris
[3]From http://archive.ics.uci.edu/ml/datasets/banknote+authentication
[4]From http://archive.ics.uci.edu/ml/datasets/Mushroom

TABLE I
PERFORMANCE (PURITY) OF DIFFERENT METHODS ON THE DATASETS D-512, IRIS AND BANKNOTE (BANK.).

|  | HC | K-means | GMM | AP | Proposed |
|---|---|---|---|---|---|
| D-512 | **1** | 0.875 | 0.874 | **1** | **1** |
| Iris | 0.893 | 0.893 | 0.667 | 0.887 | **0.953** |
| Bank. | 0.851 | 0.891 | 0.937 | 0.829 | **0.991** |

## V. CONCLUSION

In this paper, we proposed an automatic method to remove the undesired edges in the ND-based In-tree via an effective combination of In-tree with AP, so as to reveal the underlying cluster structure in the dataset. The experiments on both synthetic and real datasets demonstrated the effectiveness of the proposed method.

ACKNOWLEDGMENT

This work was supported by the 973 Project under Grant 2013CB329401, the NSFC under Grant 61375115, 91420105, and the Doctoral Support Program of UESTC.

REFERENCES

[1] C. V. Cannistraci, T. Ravasi, F. M. Montevecchi, T. Ideker, and M. Alessio. Nonlinear dimension reduction and clustering by minimum curvilinearity unfold neuropathic pain and tissue embryological classes. *Bioinformatics*, 26(18):i531–i539, 2010.
[2] H. Chang and D.-Y. Yeung. Robust path-based spectral clustering. *Pattern Recognition*, 41(1):191–203, 2008.
[3] P. Franti and O. Virmajoki. Iterative shrinking method for clustering problems. *Pattern Recognition*, 39(5):761–775, 2006.
[4] P. Franti, O. Virmajoki, and V. Hautamaki. Fast agglomerative clustering using a k-nearest neighbor graph. *IEEE Trans. Pattern Analysis and Machine Intelligence*, 28(11):1875–1881, 2006.
[5] B. J. Frey and D. Dueck. Clustering by passing messages between data points. *Science*, 315(5814):972–976, 2007.
[6] L. Fu and E. Medico. Flame, a novel fuzzy clustering method for the analysis of dna microarray data. *BMC bioinformatics*, 8(1):3, 2007.
[7] A. Gionis, H. Mannila, and P. Tsaparas. Clustering aggregation. *ACM Trans. Knowledge Discovery from Data*, 1(1):4, 2007.
[8] A. K. Jain. Data clustering: 50 years beyond k-means. *Pattern recognition letters*, 31(8):651–666, 2010.
[9] D. Lashkari and P. Golland. Convex clustering with exemplar-based models. In *Advances in neural information processing systems*, pages 825–832, 2007.
[10] M. Leone, M. Weigt, et al. Clustering by soft-constraint affinity propagation: applications to gene-expression data. *Bioinformatics*, 23(20):2708–2715, 2007.
[11] J. Macqueen. Some methods for classification and analysis of multivariate observations. In *Proc. 5th Berkeley Symposium on Mathematical Statistics and Probability*, pages 281–297, 1967.
[12] G. McLachlan and D. Peel. Finite mixture models: Wiley series in probability and mathematical statistics, 2000.
[13] T. Qiu and Y. Li. It-map: an effective nonlinear dimensionality reduction method for interactive clustering. *arXiv preprint arXiv:1501.06450*, 2015.
[14] T. Qiu, K. Yang, C. Li, and Y. Li. A physically inspired clustering algorithm: to evolve like particles. *arXiv preprint arXiv:1412.5902*, 2014.
[15] J. B. Tenenbaum, V. De Silva, and J. C. Langford. A global geometric framework for nonlinear dimensionality reduction. *Science*, 290(5500):2319–2323, 2000.
[16] J. H. Ward Jr. Hierarchical grouping to optimize an objective function. *Journal of the American statistical association*, 58(301):236–244, 1963.

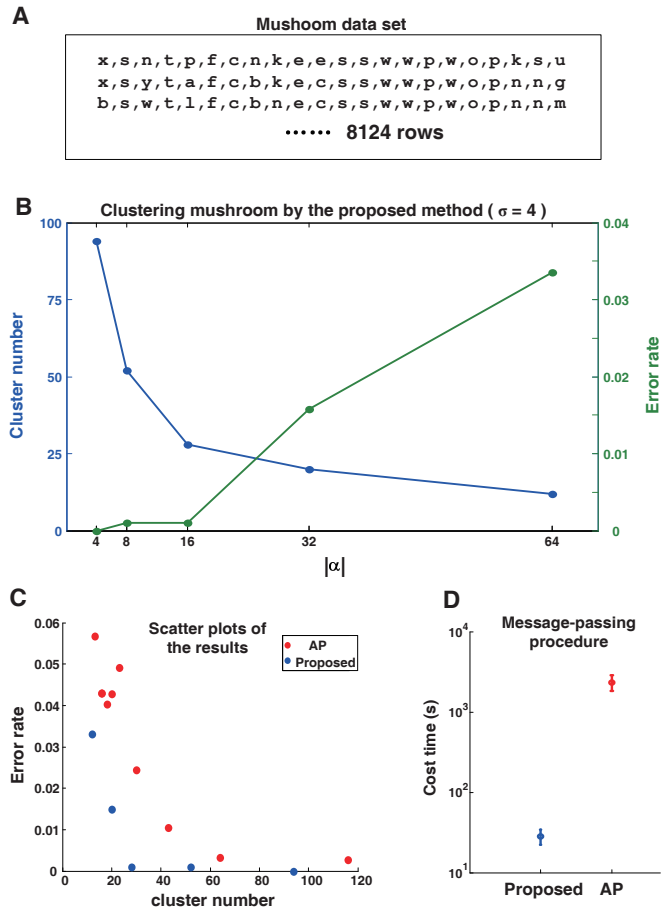

Fig. 4. (A) A small portion of mushroom dataset. (B) The clustering results (blue:cluster number; green: error rate = 1 - purity, i.e., the impurity of each cluster) achieved by the proposed method as the magnitude of the parameter $\alpha$ changes from 4 to 64. (C) Comparison between the proposed method and AP. Compared with the results of AP (red points), the results of the proposed method (blue points) are overall more close to the origin of the coordinates, indicating that the proposed method can achieve lower false clustering assignments or less cluster numbers. (D) Comparison of the time cost for the message-passing procedure between the proposed method and AP. The proposed method: $28.1 \pm 6.0$ s $(mean \pm std)$; AP: $2342.7 \pm 506.2$ s $(mean \pm std)$.

[17] R. Xu, D. Wunsch, et al. Survey of clustering algorithms. *IEEE Trans. Neural Networks*, 16(3):645–678, 2005.
[18] C. T. Zahn. Graph-theoretical methods for detecting and describing gestalt clusters. *IEEE Trans. Computers*, 100(1):68–86, 1971.